\documentclass[11pt,letterpaper]{article}

\usepackage{acl2016}
\usepackage{times}
\usepackage{latexsym}
\usepackage{amsmath,bm}
\usepackage{amsfonts}
\usepackage{mathrsfs}
\usepackage{amsbsy,amsthm,amssymb,wasysym}

\usepackage{graphicx}
\usepackage{xcolor}
\usepackage{multirow}
\usepackage{paralist}
\usepackage{subfigure}
\usepackage{fancyhdr}
\aclfinalcopy % Uncomment this line for the final submission
\def\aclpaperid{***} %  Enter the acl Paper ID here

\newcommand\imdb{{\tt IMDB}}
\newcommand\mr{{\tt MR}}
\newcommand\qc{{\tt QC}}

\newcommand\snli{{\tt SNLI}}
\newcommand\sick{{\tt SICK}}
\newcommand\msrp{{\tt MSRP}}
\newcommand\snlisick{{\tt SNLI}$\rightarrow${\tt SICK}}
\newcommand\snlimsrp{{\tt SNLI}$\rightarrow${\tt MSRP}}
\newcommand\imdbmr{{\imdb}$\rightarrow${\mr}}
\newcommand\imdbqc{{\imdb}$\rightarrow${\qc}}

\newcommand\lock{\includegraphics{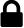}}
\newcommand\myheart{\includegraphics[scale=.8]{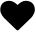}}
\newcommand\unlock{\includegraphics{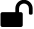}}
\newcommand\xbox{$\XBox$}

\newcommand\boxunlock{\makebox[.43cm][l]{\includegraphics{unlock.pdf}}}
\newcommand\boxlock{\makebox[.43cm][l]{\includegraphics{lock.pdf}}}
\newcommand\Boxbox{\makebox[.43cm][l]{$\Box$}}
\newcommand\boxxbox{\makebox[.43cm][l]{$\XBox$}}

\title{How Transferable are Neural Networks in NLP Applications?}
\author{Lili Mou,$^{1}$ Zhao Meng,$^1$ Rui Yan,$^2$ Ge Li,$^{1,\dag}$ Yan Xu,$^{1,}$\thanks{\ \ Yan Xu is currently a research scientist at Inveno Co., Ltd.}\ \ \ Lu Zhang,$^{1}$ Zhi Jin$^{1,\dag}$\\
$^{1}$Key Laboratory of High Confidence Software Technologies (Peking University), MoE, China\\
Institute of Software, Peking University, China\quad $^\dag$Corresponding authors\\
$^{2}$Insitute of Computer Science and Technology of Peking University, China\\ \normalsize{\tt \{doublepower.mou,rui.yan.peking\}@gmail.com,zhaomeng.pku@outlook.com}\\
\normalsize{\tt \{lige,xuyan14,zhanglu,zhijin\}@sei.pku.edu.cn}\\
}

\date{}

\begin{document}

\maketitle
\renewcommand{\headrulewidth}{0pt}
\renewcommand{\footskip}{30pt}
\cfoot{Accepted by EMNLP-2016}
\thispagestyle{fancy}
\begin{abstract}
Transfer learning is aimed to make use of valuable knowledge in a \textit{source} domain to help model performance in a \textit{target} domain. It is particularly important to neural networks, which are very likely to be overfitting. In some fields like image processing, many studies have shown the effectiveness of neural network-based transfer learning. For neural NLP, however, existing studies have only casually applied transfer learning, and conclusions are inconsistent. In this paper, we conduct systematic case studies and provide an illuminating picture on the transferability of neural networks in NLP.\footnote{Code released on https://sites.google.com/site/transfernlp/}
\end{abstract}

\section{Introduction}\label{sec:intro}

Transfer learning, or sometimes known as domain adaptation,\footnote{In this paper, we do not distinguish the conceptual difference between \textit{transfer learning} and \textit{domain adaptation}. \textit{Domain}---in the sense we use throughout this paper---is defined by datasets.} plays an important role in various natural language processing (NLP) applications, especially when we do not have large enough datasets for the task of interest (called the \textit{target} task $\mathcal{T}$). In such scenarios, we would like to transfer or adapt knowledge from other domains (called the \textit{source} domains/tasks $\mathcal{S}$) so as to mitigate the problem of overfitting and to improve model performance in $\mathcal{T}$. For traditional feature-rich or kernel-based models, researchers have developed a variety of elegant methods for domain adaptation; examples include EasyAdapt \cite{easyAdapt,easyAdapt2}, instance weighting \cite{instanceWeighting,instanceWeighting2}, and structural correspondence learning \cite{SCL,SCL2}.

Recently, deep neural networks are emerging as the prevailing technical solution to almost every field in NLP. Although capable of learning highly nonlinear features, deep neural networks are very prone to overfitting, %\cite{regularization} 
compared with traditional methods. Transfer learning therefore becomes even more important. Fortunately, neural networks can be trained in a transferable way by their incremental learning nature: we can directly use trained (tuned) parameters from a source task to initialize the network in the target task; alternatively, we may also train two tasks simultaneously with some parameters shared. But their performance should be verified by empirical experiments.

Existing studies have already shown some evidence of the transferability of neural features. For example, in image processing, low-level neural layers closely resemble Gabor filters or color blobs \cite{visualize,filterblob}; they can be transferred well to different tasks. \newcite{image} suggest that high-level layers are also transferable in general visual recognition; \newcite{how} further investigate the transferability of neural layers in different levels of abstraction.

Although transfer learning is promising in image processing, conclusions appear to be less clear in NLP applications. Image pixels are low-level signals, which are generally continuous and less related to semantics. By contrast, natural language tokens are discrete: each word well reflects the thought of humans, but neighboring words do not share as much information as pixels in images do. 
Previous neural NLP studies have casually applied transfer techniques, but their results are not consistent. \newcite{unified} apply multi-task learning to SRL, NER, POS, and CHK,\footnote{
The acronyms refer to \textit{semantic role labeling}, \textit{named entity recognition}, \textit{part-of-speech tagging}, and \textit{chunking}, respectively.} but obtain only 0.04--0.21\% error reduction\footnote{Here, we quote the accuracies obtained by using unsupervised pretraining of word embeddings. This is the highest performance in that paper; using pretrained word embeddings is also a common practice in the literature.% Besides, we would like to emphasize that our paper mainly focuses on transferring knowledge from supervised tasks.
} (out of a base error rate of 16--18\%).
\newcite{NLI}, on the contrary, improve a natural language inference task from an accuracy of 71.3\% to 80.8\% by initializing parameters with an additional dataset of 550,000 samples. %Other studies of neural network-based transfer learning include \newcite{adaptsentiment}, \newcite{multDRR}, etc.
Therefore, more systematic studies are needed to shed light on transferring neural networks in the field of NLP.

\subsection*{Our Contributions}
In this paper, we investigate the question ``\textit{How transferable are neural networks in NLP applications?}'' 

We distinguish two scenarios of transfer: (1) transferring knowledge to a semantically similar/equivalent task but with a different dataset; (2) transferring knowledge to a task that is semantically different but shares the same neural topology/architecture so that neural parameters can indeed be transferred. We further distinguish two transfer methods: (1) using the parameters trained on $\mathcal{S}$ to initialize $\mathcal{T}$ (INIT), and (2) multi-task learning (MULT), i.e., training $\mathcal{S}$ and $\mathcal{T}$ simultaneously. (Please see Sections~\ref{sec:dataset} and \ref{sec:transfer}). Our study mainly focuses on the following research questions:

\begin{itemize}
\item[\textbf{RQ1:}] How transferable are neural networks between two tasks with similar or different semantics in NLP applications? 
\item[\textbf{RQ2:}] How transferable are different layers of NLP neural models?
\item[\textbf{RQ3:}] How transferable are INIT and MULT, respectively? What is the effect of combining these two methods?
\end{itemize}

\smallskip
We conducted extensive experiments over six datasets on classifying sentences and sentence pairs. We leveraged the widely-used convolutional neural network (CNN) and long short term memory (LSTM)-based recurrent neural network (RNN) as our models. 

Based on our experimental results, we have the following main observations, some of which are unexpected.

\medskip
\noindent\fbox{
\parbox{.45\textwidth}{
\begin{compactitem}[$\bullet$]
\item Whether a neural network is transferable in NLP depends largely on how  semantically similar the tasks are, which is different from the consensus in image processing.

\item The output layer is mainly specific to the dataset and not transferable. Word embeddings are likely to be transferable to semantically different tasks.

\item MULT and INIT appear to be generally comparable to each other; combining these two methods does not result in further gain in our study.
\end{compactitem}
}
}

\bigskip

The rest of this paper is organized as follows. Section~\ref{sec:dataset} introduces the datasets that neural models are transferred across; Section~\ref{sec:setting} details the neural architectures and experimental settings. We describe two approaches (INIT and MULT) to transfer learning in Section~\ref{sec:transfer}. We present experimental results in Sections~\ref{sec:INIT}--\ref{sec:MULT} and have concluding remarks in Section~\ref{sec:conclusion}. 

\bigskip
\section{Datasets}\label{sec:dataset}
\begin{table}[!t]
\centering
\resizebox{.49\textwidth}{!}{
\begin{tabular}{|l||r|r|r||r|r|r|}
\hline
\multicolumn{7}{|c|}{\textbf{Statistics (\# of Samples)}}\\
\hline   & \multicolumn{3}{|c||}{\textbf{Experiment I}} &
          \multicolumn{3}{|c|}{\textbf{Experiment II}}\\
\hline
         & \imdb & \mr & \qc & \snli & \sick & \msrp\\
\hline\hline
\#Train\!\! & 550,000 & 8,500 & 4,800 & 550,152 & 4,439 & 3,575\\ 
\#Val   &  50,000 & 1,100 & 600  & 10,000  & 495   & 501  \\
\#Test  &   2,000  & 1,100 & 500  & 10,000  & 4,906 & 1,725\\
\hline
\end{tabular}
}
\resizebox{.49\textwidth}{!}{
\begin{tabular}{|l|l|l|}
\hline
\multicolumn{3}{|c|}{\textbf{Examples in Experiment~I}} \\
\hline
\multicolumn{3}{|c|}{Sentiment Analysis (\imdb\ and \mr)}\\
\hline
\multicolumn{2}{|l|}{An idealistic love story that brings out }& \multirow{2}{*}{$+$}\\
\multicolumn{2}{|l|}{the latent 15-year-old romantic in everyone.}&  \\
\hline
\multicolumn{2}{|l|}{Its mysteries are transparently obvious,} & \multirow{2}{*}{$-$}\\
\multicolumn{2}{|l|}{and it’s too slowly paced to be a thriller.} &\\
\hline\hline
\multicolumn{3}{|c|}{Question Classification (\qc)}\\\hline
\multicolumn{2}{|l|}{What is the temperature at the center of the earth?}  & \texttt{number}\\
\multicolumn{2}{|l|}{What state did the Battle of Bighorn take place in?} & \texttt{location}\\
\hline
\hline\multicolumn{3}{|c|}{\textbf{Examples in Experiment~II}} \\
\hline
\multicolumn{3}{|c|}{Natural Language Inference (\snli\ and \sick)}\\
\hline
\textbf{Premise} & Two men on bicycles competing in a race. & \\
\hline
                    & People are riding bikes. & {\tt E}\\
\textbf{Hypothesis} & Men are riding bicycles on the streets. & {\tt C}\\
                    & A few people are catching fish. & {\tt N}\\
\hline\hline
%Two men on bicycles competing in a race. & People are riding bikes. & {\tt Entailment}\\
%Men are riding bicycles on the streets. &{\tt Contradiction}\\
% A few people are catching fish. & {\tt Neutral} \\
\multicolumn{3}{|c|}{Paraphrase Detection (\msrp)}\\
\hline
 \multicolumn{2}{|l|}{
The DVD-CCA then appealed to the state} & \multirow{4}{*}{\text{Paraphrase}}\\

 \multicolumn{2}{|l|}{Supreme Court.} & \\\cline{1-2}
                    \multicolumn{2}{|l|}{The DVD CCA appealed that decision} & \\
                   \multicolumn{2}{|l|}{to the U.S. Supreme Court.} & \\
\hline
       \multicolumn{2}{|l|}{Earnings per share from recurring operations} & \\
      \multicolumn{2}{|l|}{will be 13 cents to 14 cents.} & \text{Non-} \\
                    \cline{1-2}
 \multicolumn{2}{|l|}{That beat the company's April earnings } &\text{Paraphrase}\\
 \multicolumn{2}{|l|}{forecast of 8 to 9 cents a share.} &\\
\hline
\end{tabular}
}
\caption{Statistics and examples of the datasets.}
\label{tab:dataset}
\end{table}
In our study, we conducted two series of experiments using six open datasets as follows.
\begin{itemize}[$\bullet$]
\item \textbf{Experiment~I}: Sentence classification
\begin{compactitem}[$-$]%\small
\item \imdb. A large dataset for binary sentiment classification (\texttt{positive} vs. \texttt{negative}).\footnote{https://drive.google.com/file/d/\\
{\color{white}}\quad\quad\quad 0B8yp1gOBCztyN0JaMDVoeXhHWm8/}
\item \mr. A small dataset for binary sentiment classification.\footnote{https://www.cs.cornell.edu/people/pabo/\\
{\color{white}}\quad\quad\quad movie-review-data/}
\item \qc. A (small) dataset for 6-way question classification (e.g., \texttt{location}, \texttt{time}, and \texttt{number}).\footnote{http://cogcomp.cs.illinois.edu/Data/QA/QC/}
\end{compactitem}
\newpage
\item \textbf{Experiment~II}: Sentence-pair classification
\begin{compactitem}[$-$]%\small
\item \snli. A large dataset for sentence entailment recognition. The classification objectives are \texttt{entailment}, \texttt{contradiction}, and \texttt{neutral}.\footnote{http://nlp.stanford.edu/projects/snli/}
\item \sick. A small dataset with exactly the same classification objective as SNLI.\footnote{http://http://alt.qcri.org/semeval2014/task1/}
\item \msrp. A (small) dataset for paraphrase detection. The objective is binary classification: judging whether two sentences have the same meaning.\footnote{http://research.microsoft.com/en-us/downloads/}
\end{compactitem}

\end{itemize}
In each experiment, the large dataset serves as the source domain and small ones are the target domains. 
Table~\ref{tab:dataset} presents statistics of the above datasets.

We distinguish two scenarios of transfer regarding semantic similarity: (1) semantically equivalent transfer (\imdbmr, \snlisick), that is,  the tasks of $\mathcal S$ and $\mathcal T$ are defined by the same meaning, and (2) semantically different transfer (\imdbqc, \snlimsrp). Examples are also illustrated in Table~\ref{tab:dataset} to demonstrate semantic relatedness.

It should be noticed that in image or speech processing \cite{how,speech}, the input of neural networks pretty much consists of raw signals; hence, low-level feature detectors are almost always transferable, even if \newcite{how} manually distinguish artificial objects and natural ones in an image classification task.  

Distinguishing semantic relatedness---which emerges from very low layers of either word embeddings or the successive hidden layer---is specific to NLP and also a new insight of our paper. As we shall see in Sections~\ref{sec:INIT} and~\ref{sec:MULT}, the transferability of neural networks in NLP is more sensitive to semantics than in image processing.

\section{Neural Models and Settings}\label{sec:setting}

\begin{figure}[!t]
\centering
\includegraphics[width=.85\linewidth]{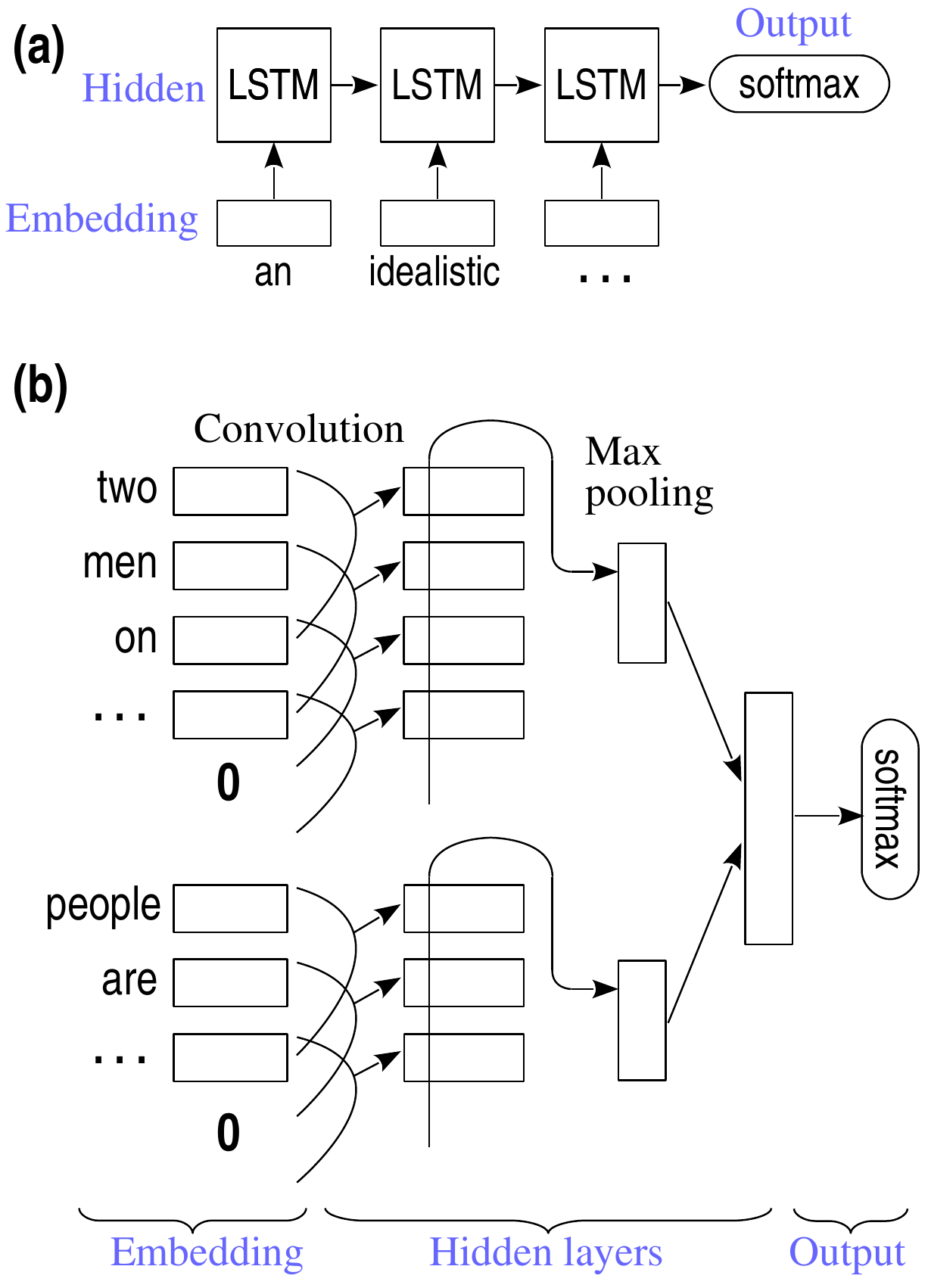}
\caption{The models in our study. (a) Experiment I: RNNs with LSTM units for sentence classification. (b) Experiment II: CNN for sentence pair modeling.}\label{fig:model}
\end{figure}

In each group, we used a single neural model to solve three problems in a unified manner.
That is to say, the neural architecture is the same among the three datasets, which makes it possible to investigate transfer learning regardless of whether the tasks are semantically equivalent. Concretely, the neural models are as follows. %(See also Appendix B.)
\begin{itemize}
\item\textbf{Experiment~I: LSTM-RNN}. To classify a sentence according to its sentiment or question type, we use a recurrent neural network (RNN, Figure~\ref{fig:model}a) with long short term memory (LSTM) units~\cite{lstm}. A softmax layer is added to the last word's hidden state for classification. 
\item\textbf{Experiment~II: CNN-pair}. In this group, we use a ``Siamese'' architecture \cite{siamese} to classify the relation of two sentences. We first apply a convolutional neural network (CNN, Figure~\ref{fig:model}b) with a window size of 5 to model local context, and a max pooling layer gathers information to a fixed-size vector. Then the sentence vectors are concatenated and fed to a hidden layer before the softmax output.
\end{itemize}

%\begin{figure}[!t]
%\centering
%\includegraphics[width=.4\textwidth]{model.pdf}
%\caption{The neural model in our study.}
%\label{fig:model}
%\end{figure}

In our experiments, embeddings were pretrained by {\tt word2vec} \cite{word2vec}; all embeddings and hidden layers were 100 dimensional. 
We applied stochastic gradient descent with a mini-batch size of 50 for optimization. In each setting, we tuned the hyperparameters as follows: learning rate from $\{3, 1, 0.3, 0.1,0.03\}$, power decay of learning rate from $\{$fast, moderate, low$\}$ (defined by how much, after one epoch, the learning rate residual is: $0.1\text{x}, 0.3\text{x}, 0.9\text{x},$ resp). We regularized our network by dropout with a rate from $\{0,0.1,0.2,0.3\}$. Note that we might not run nonsensical settings, e.g., a larger dropout rate if the network has already been underfitting (i.e., accuracy has decreased when the dropout rate increases). %However, we might try additional smaller learning rates, if needed, especially in transfer learning by INIT.
We report the test performance associated with the highest validation accuracy.

To setup a baseline, we trained our models without transfer 5 times by different random parameter initializations (Table~\ref{tab:notransfer}).
We have achieved reasonable performance that is comparable to similar models reported in the literature with all six datasets. Therefore, our implementation is fair and suitable for further study of transfer learning. 

It should be mentioned that the goal of this paper is not to outperform state-of-the-art results; instead, we would like to conduct a fair comparison of different methods and settings for transfer learning in NLP.

\begin{table}[!t]
\centering
\resizebox{.48\textwidth}{!}{
\begin{tabular}{|cccl|}
\hline
\multicolumn{2}{|c}{\textbf{Dataset}}&\!\!\!\! \textbf{Avg acc.$\pm$std.}\!\!\!\!\!\! &\multicolumn{1}{c|}{\textbf{Related model}}\\\hline\hline
\multirow{3}{*}{\rotatebox{90}{\textbf{Exp.~I}}} & \!\!\!\!\imdb & 87.0              & 89.3 (Non-NN, Dong$^+$,2015)\!\!\! \\
  &\!\!\!\!\mr   & $75.1\pm 0.6$     & 77.7 (RAE, Socher$^+$, 2013)\\
  &\!\!\!\!\qc   & $90.8 \pm 0.9$    & 90.2 (RNN, Zhao$^+$,2015)\\
\hline
\multirow{3}{*}{\rotatebox{90}{\!\textbf{Exp.~II}}}   & \!\!\!\!\snli  & $76.3$     	   &  77.6 (RNN, Bowman$^+$,2015)\\
  &\!\!\!\!\sick  & $70.9\pm 1.3$     &  71.3 (RNN, Bowman$^+$,2015)\\
  &\!\!\!\!\msrp  & $69.0\pm 0.5$     &  69.6 (Arc-I CNN, Hu$^+$,2014)\\
\hline 
\end{tabular}
}
\caption{Accuracy (\%) without transfer. We also include related models for comparison \protect\cite{CL,recursive,self-adaptive,NLI,CNN:NIPS}, showing that we have achieved comparable results, and thus are ready to investigate transfer learning. The models were run one only once in source domains, because we could only transfer a particular model instead of an average of several models.}\label{tab:notransfer}
\end{table}

\section{Transfer Methods}\label{sec:transfer}

Transfer learning aims to use knowledge in a source domain to aid the target domain. As neural networks are usually trained incrementally with gradient descent (or variants), it is straightforward to use gradient information in both source and target domains for optimization so as to accomplish knowledge transfer. Depending on how samples in source and target domains are scheduled, there are two main approaches to neural network-based transfer learning:

\begin{itemize}[$\bullet$]
\item Parameter initialization (INIT). The INIT approach first trains the network on $\mathcal{S}$, and then directly uses the tuned parameters to initialize the network for $\mathcal{T}$. After transfer, we may fix~(\lock) the parameters in the target domain \cite{adaptsentiment}, i.e., no training is performed on $\mathcal{T}$. But when labeled data are available in $\mathcal{T}$, it would be better to fine-tune~(\unlock) the parameters.

INIT is also related to unsupervised pretraining such as word embedding learning \cite{word2vec} and autoencoders \cite{autoencoder}. In these approaches, parameters that are (pre)trained in an unsupervised way are transferred to initialize the model for a supervised task \cite{funnypivotcorpus}. However, our paper focuses on ``supervised pretraining,'' which means we transfer knowledge from a labeled source domain.
\item Multi-task learning (MULT). MULT, on the other hand, simultaneously trains samples in both domains \cite{unified,multDRR}. The overall cost function is given by
\begin{equation}
J=\lambda J_\mathcal{T}+(1-\lambda)J_\mathcal{S}\label{eqn:J}
\end{equation}
where $J_\mathcal{T}$ and $J_\mathcal{S}$ are the individual cost function of each domain. (Both $J_\mathcal{T}$ and $J_\mathcal{S}$ are normalized by the number of training samples.)
$\lambda\in(0,1)$ is a hyperparameter balancing the two domains.

It is nontrivial to optimize Equation~\ref{eqn:J} in practice by gradient-based methods. One may take the partial derivative of $J$ and thus $\lambda$ goes to the learning rate \cite{multDRR}, but the model is then vulnerable because it is likely to blow up with large learning rates (multiplied by $\lambda$ or $1-\lambda$) and be stuck in local optima with small ones.

\newcite{unified} alternatively choose a data sample from either domain with a certain probability (controlled by $\lambda$) and take the derivative for the particular data sample. In this way, domain transfer is independent of learning rates, but we may not be able to fully use the entire dataset of $\mathcal{S}$ if $\lambda$ is large. We adopted the latter approach in our experiment for simplicity. (More in-depth analysis may be needed in future work.) Formally, our multi-task learning strategy is as follows.
\begin{compactitem}
\item[1] Switch to $\mathcal{T}$ with prob.~$\lambda$, or to $\mathcal{S}$ with prob.~$1-\lambda$.
\item[2] Compute the gradient of the next data sample in the particular domain.
\end{compactitem}
\end{itemize}

Further, INIT and MULT can be combined straightforwardly, and we obtain the third setting:

\begin{itemize}[$\bullet$]
\item Combination (MULT+INIT). We first pretrain on the source domain $\mathcal{S}$ for parameter initialization, and then train $\mathcal{S}$ and $\mathcal{T}$ simultaneously.
\end{itemize}

\smallskip
From a theoretical perspective, INIT and MULT work in different ways. In the MULT approach, the source domain regularizes the model by ``aliasing'' the error surface of the target domain; hence the neural network is less prone to overfitting. In INIT, $\mathcal{T}$'s error surface remains intact. Before training on the target dataset, the parameters are initialized in such a meaningful way that they contain additional knowledge in the source domain. However, in an extreme case where $\mathcal{T}$'s error surface is convex, INIT is ineffective because the parameters can reach the global optimum regardless of their initialization. In practice, deep neural networks usually have highly complicated, non-convex error surfaces. By properly initializing parameters with the knowledge of $\mathcal{S}$, we can reasonably expect that the parameters are in a better ``catchment basin,'' and that the INIT approach can transfer knowledge from $\mathcal{S}$ to $\mathcal{T}$.

\section{Results of Transferring by INIT}\label{sec:INIT}

We first analyze how INIT behaves in NLP-based transfer learning. In addition to two different transfer scenarios regarding semantic relatedness as described in Section~\ref{sec:dataset}, we further evaluated two settings: (1) fine-tuning parameters \unlock, and (2) freezing parameters after transfer \lock. Existing evidence shows that frozen parameters would generally hurt the performance~\cite{regularization}, but this setting provides a more direct understanding on how transferable the features are (because the factor of target domain optimization is ruled out). Therefore, we included it in our experiments. Moreover, we transferred parameters layer by layer to answer our second research question. %Notice that the E\lock H\lock O\lock\ and E\unlock  H\unlock O\unlock\ settings\footnote{Please refer to the caption of Table~\ref{tab:INIT} for the \xbox, $\Box$, \unlock, and \lock\ symbols.} are inapplicable to \imdbqc\ and \snlimsrp, because the output targets do not share same meanings and numbers of target classes.

Through Subsections~\ref{ss:performance}--\ref{ss:learningRate}, we initialized the parameters of $\mathcal{T}$ with the ones corresponding to the highest validation accuracy of $\mathcal{S}$. In Subsection~\ref{ss:when}, we further investigated when the parameters are ready to be transferred during the training on $\mathcal{S}$.

\subsection{Overall Performance}\label{ss:performance}
Table~\ref{tab:INIT} shows the main results of INIT. A quick observation is that, in both groups, transfer learning of semantically equivalent tasks (\imdbmr, \snlisick) appears to be successful with an improvement of $\sim$6\%. The results are not surprising and also reported in \newcite{NLI}.  

For \imdbqc\ and \snlimsrp, however, there is no improvement of transferring hidden layers (embeddings excluded), namely LSTM-RNN units and CNN feature maps. The E\unlock H\unlock O$\Box$ setting yields a slight degradation of 0.2--0.4\%, $\sim$.5x std. The incapability of transferring is also proved by locking embeddings and hidden layers (E\lock H\lock O$\Box$). We see in this setting, the test performance is very low in \qc\ or even worse than majority-class guess in \msrp. By further examining its training accuracy, which is 48.2\% and 65.5\%, respectively, we conclude that extracted features by LSTM-RNN and CNN models in $\mathcal{S}$ are almost irrelevant to the ultimate tasks $\mathcal{T}$ (\qc\ and \msrp).

Although in previous studies, researchers have mainly drawn positive conclusions about transfer learning, we find a negative result similar to ours upon careful examination of \newcite{unified}, and unfortunately, their results may be somewhat misinterpreted. In that paper, the authors report transferring NER, POS, CHK, and pretrained word embeddings improves the SRL task by 1.91--3.90\% accuracy (out of 16.54--18.40\% error rate), but their gain is mainly due to word embeddings. In the settings that use pretrained word embeddings (which is common in NLP), NER, POS, and CHK together improve the SRL accuracy by only 0.04--0.21\%.

The above results are rather frustrating, indicating for RQ1 that neural networks may not be transferable to NLP tasks of different semantics. Transfer learning for NLP is more prone to semantics than the image processing domain, where even high-level feature detectors are almost always transferable \cite{image,how}.

\subsection{Layer-by-Layer Analysis}
\begin{table}[!t]
\centering
\resizebox{.4\textwidth}{!}{
\begin{tabular}{|c|r|r|}
\hline
\multicolumn{3}{|c|}{\textbf{Experiment I}}\\
\hline
Setting &\!\!\! \imdbmr \!\!&\!\!\! \imdbqc\!\!\\
\hline\hline
Majority                         &    50.0  & 22.9\!\!\\
\!\!E\boxxbox    H\Boxbox    O\Boxbox\!\!\!\! & 75.1 & 90.8\!\!\\
\hline
\!\!E\boxlock   H\Boxbox    O\Boxbox\!\!\!\! & 78.2 & 93.2\!\!\\
\!\!E\boxlock  H\boxlock   O\Boxbox\!\!\!\!  & 78.8 & 55.6\!\!\\
\!\!E\boxlock  H\boxlock   O\boxlock\!\!\!\! & 73.6 & --\!\!\\
\hline
\!\!E\boxunlock   H\Boxbox    O\Boxbox\!\!\!\!     & 78.3 & 92.6\!\!\\
\!\!E\boxunlock  H\boxunlock   O\Boxbox\!\!\!\!    & 81.4 & 90.4\!\!\\
\!\!E\boxunlock  H\boxunlock   O\boxunlock\!\!\!\! & 80.9 & -- \!\!\\
\hline\hline
\multicolumn{3}{|c|}{\textbf{Experiment II}}\\
\hline
Setting &\!\!\! \snlisick \!\!&\!\!\! \snlimsrp\!\!\\
\hline\hline
Majority                         &   56.9   & 66.5\!\!\\
\!\!E\boxxbox    H\Boxbox    O\Boxbox\!\!\!\! & 70.9 & 69.0\!\!\\
\hline
\!\!E\boxlock   H\Boxbox    O\Boxbox\!\!\!\! & 69.3 & 68.1\!\!\\
\!\!E\boxlock  H\boxlock   O\Boxbox\!\!\!\!  & 70.0 & 66.4\!\!\\
\!\!E\boxlock  H\boxlock   O\boxlock\!\!\!\! &43.1 & --\!\!\\
\hline
\!\!E\boxunlock   H\Boxbox    O\Boxbox\!\!\!\!     & 71.0 & 69.9\!\!\\
\!\!E\boxunlock  H\boxunlock   O\Boxbox\!\!\!\!    & 76.3 & 68.8\!\!\\
\!\!E\boxunlock  H\boxunlock   O\boxunlock\!\!\!\! & 77.6 & -- \!\!\\
\hline
\end{tabular}
}
\caption{Main results of neural transfer learning by INIT. We report test accuracies (\%) in this table. E: embedding layer; H: hidden layers; O: output layer. $\XBox$: Word embeddings are pretrained by {\tt word2vec};
$\Box$: Parameters are randomly initialized); \lock: Parameters are transferred but frozen;
\unlock: Parameters are transferred and fine-tuned. Notice that the E\lock H\lock O\lock\ and E\unlock  H\unlock O\unlock\ settings are inapplicable to \imdbqc\ and \snlimsrp, because the output targets do not share same meanings and numbers of target classes.} \label{tab:INIT}
\vspace{-.4cm}
\end{table}

%\begin{table}[!t]
%\centering
%\begin{tabular}{cc|cc}
%\hline
%       \multicolumn{2}{c|}{\snli}     &     \multicolumn{2}{c}{\snlimsrp}\\
%\hline
%        &           & E\Boxbox H\Boxbox O\Boxbox          &  60.3\\
%        &           & E\boxxbox H\Boxbox O\Boxbox         &  69.0\\
%E$\Box$ & 76.4      & E\boxunlock H\Boxbox O\Boxbox       &  67.2\\
%E\xbox  & 76.3      & E\boxunlock H\Boxbox O\Boxbox       &  69.9\\
%\hline   
%\end{tabular}
%\caption{Analyzing the ``supervised'' and ``unsupervised'' pretraining of word embeddings. The E\unlock\ setting refers to transferring word embeddings from \snli. However, \snli's embeddings themselves are either randomly initialized or pretrained by {\tt word2vec}, indicating by E$\Box$ and E\xbox\ in the first column.}\label{tab:embed}
%\end{table}

To answer RQ2, we next analyze the transferability of each layer. 
First, we freeze both embeddings and hidden layers (E\lock H\lock). Even in semantically equivalent settings, if we further freeze the output layer (O\lock), the performance in both \imdbmr\ and \snlisick\ drops, but by randomly initializing the output layer's parameters (O$\Box$), we can obtain a similar or higher result compared with the baseline (E\xbox H$\Box$O$\Box$).
The finding suggests that the output layer is mainly specific to a dataset. Transferring the output layer's parameters yields little (if any) gain.

Regarding embeddings and hidden layers (in the settings E\unlock{}H\unlock O$\Box/$E\unlock{}H$\Box$O$\Box$ vs.~E\XBox H$\Box$O$\Box$), the \imdbmr{} experiment suggests both of embeddings and the hidden layer play an important role, each improving the accuracy by 3\%. In \snlisick, however, the main improvement lies in the hidden layer. A plausible explanation is that in sentiment classification tasks (\imdb\ and \mr), information emerges from raw input, i.e., sentiment lexicons and  thus their embeddings, but natural language inference tasks (\snli\ and \sick) address more on semantic compositionality and thus hidden layers are more important.

Moreover, for semantically different tasks (\imdbqc\ and \snlimsrp), the embeddings are the only parameters that have been observed to be transferable, slightly benefiting the target task by 2.7x and 1.8x std, respectively.
%As the above studies use {\tt word2vec} to pretrain word embeddings (E\xbox) on the Wikipedia corpus (which is a common practice in the literature), a curious question is whether word embeddings are transferable, provided that they are randomly initialized.

%To answer this question, we first trained \snli\ again with word embeddings being %randomly initialized, and then transferred the pretrained word embeddings (in a %supervised manner on \snli) to \msrp. Table~\ref{tab:embed} compares the result of %transferring word embeddings (pretrained by supervised/unsupervied approaches) with %non-transferring. As is seen, random initialization of word embeddings works fine %for large datasets like \snli, but performs very poorly in \msrp.
%Transferring word embeddings that are pretrained purely by supervised objective on %\snli\ significantly improves the performance, but is still worse than {\tt %word2vec} pretraining by 1.8x std.
%Therefore, we conclude

%\medskip
%\noindent\fbox{
%\parbox{.45\textwidth}{
%Word embeddings (purely pretrained by supervised approaches on $\mathcal{S}$) are indeed transferable across semantically different tasks. When large unlabeled corpora are available, however, unsupervised pretraining of word embeddings like {\tt word2vec} is preferable.
%}
%}

% MSRP: embedding random initialization: 60.3\%
\subsection{How does learning rate affect transfer?}\label{ss:learningRate}
\begin{figure}[!t]
\centering
\textbf{\quad Experiment~I}\\
\includegraphics[width=.4\textwidth]{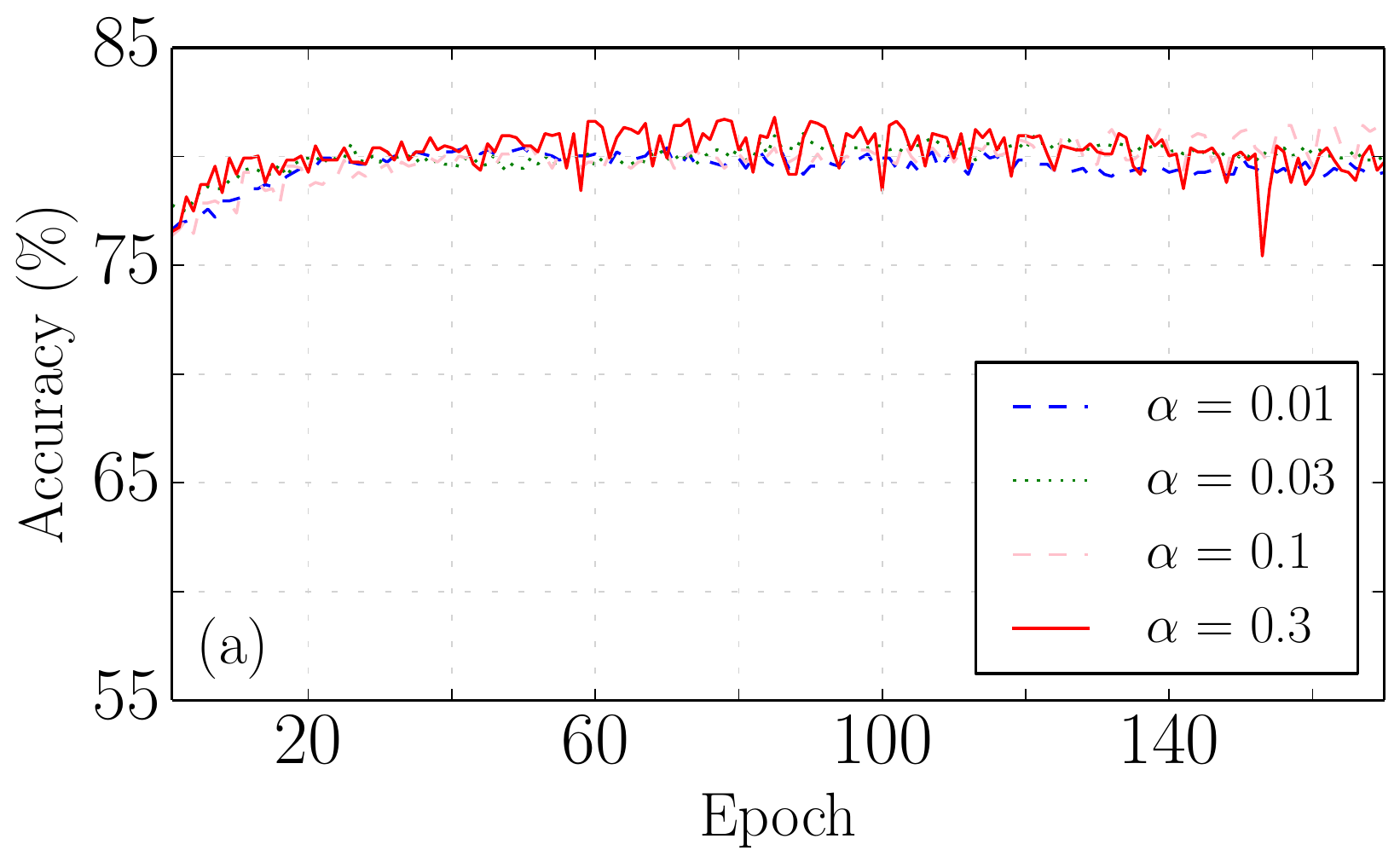}\\
\textbf{\quad Experiment~II}\\
\includegraphics[width=.4\textwidth]{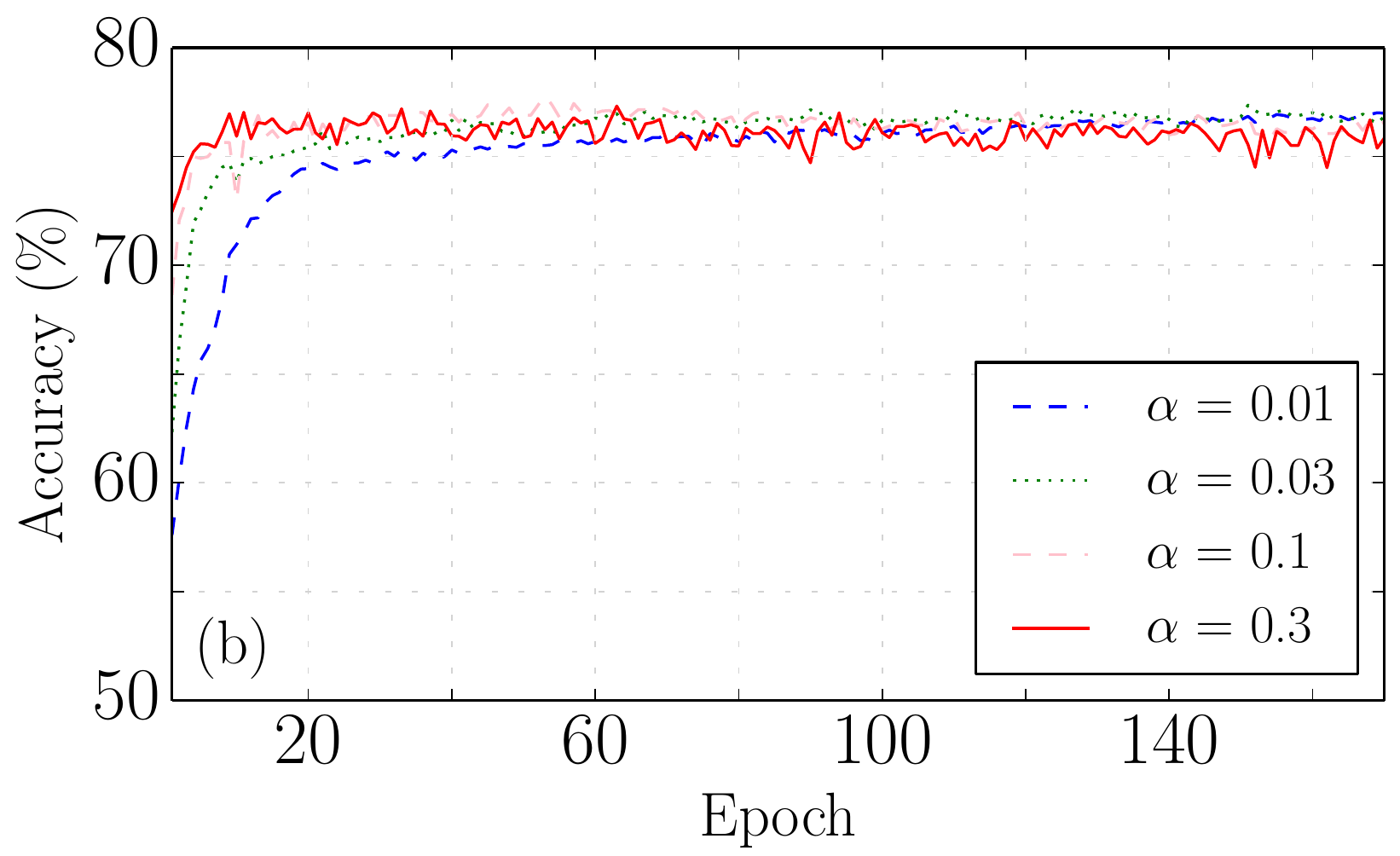}
\vspace{-.2cm}
\caption{Learning curves of different learning rates (denoted as $\alpha$). (a) Experiment~I: \imdbmr; (b) Experiment~II: \snlisick.}
\vspace{-.5cm}
\label{fig:alpha}
\end{figure}
\newcite{NLI} suggest that after transferring, a large learning rate may damage the knowledge stored in the parameters; in their paper, they transfer the learning rate information (AdaDelta) from $\mathcal{S}$ to $\mathcal{T}$ in addition to the parameters.

Although the rule of the thumb is to choose all hyperparameters---including the learning rate---by validation, we are curious whether the above conjecture holds. Estimating a rough range of sensible hyperparameters can ease the burden of model selection; it also provides evidence to better understand how transfer learning actually works.

We plot the learning curves of different learning rates $\alpha$ in Figure~\ref{fig:alpha} (\imdbmr\ and \snlisick, E\unlock H\unlock O$\Box$). (In the figure, no learning rate decay is applied.) As we see, with a large learning rate like $\alpha=0.3$, the accuracy increases fast and peaks at earlier epochs. Training with a small learning rate (e.g., $\alpha=0.01$) is slow, but its peak performance is comparable to large learning rates when iterated by, say, 100 epochs. The learning curves in Figure~\ref{fig:alpha} are similar to classic speed/variance trade-off, and we have the following additional discovery:

\smallskip
\noindent\fbox{
\parbox{.45\textwidth}{
In INIT, transferring learning rate information is not necessarily useful. A large learning rate does not damage the knowledge stored in the pretrained hyperparameters, but accelerates the training process to a large extent. In all, we may need to perform validation to choose the learning rate if computational resources are available.
}
}

\medskip
\subsection{When is it ready to transfer?}\label{ss:when}

In the above experiments, we transfer the parameters when they achieve the highest validation performance on $\mathcal{S}$. This is a straightforward and intuitive practice. 

\begin{figure}[!t]
\centering
\textbf{\quad Experiment~I}\\
\includegraphics[width=.42\textwidth]{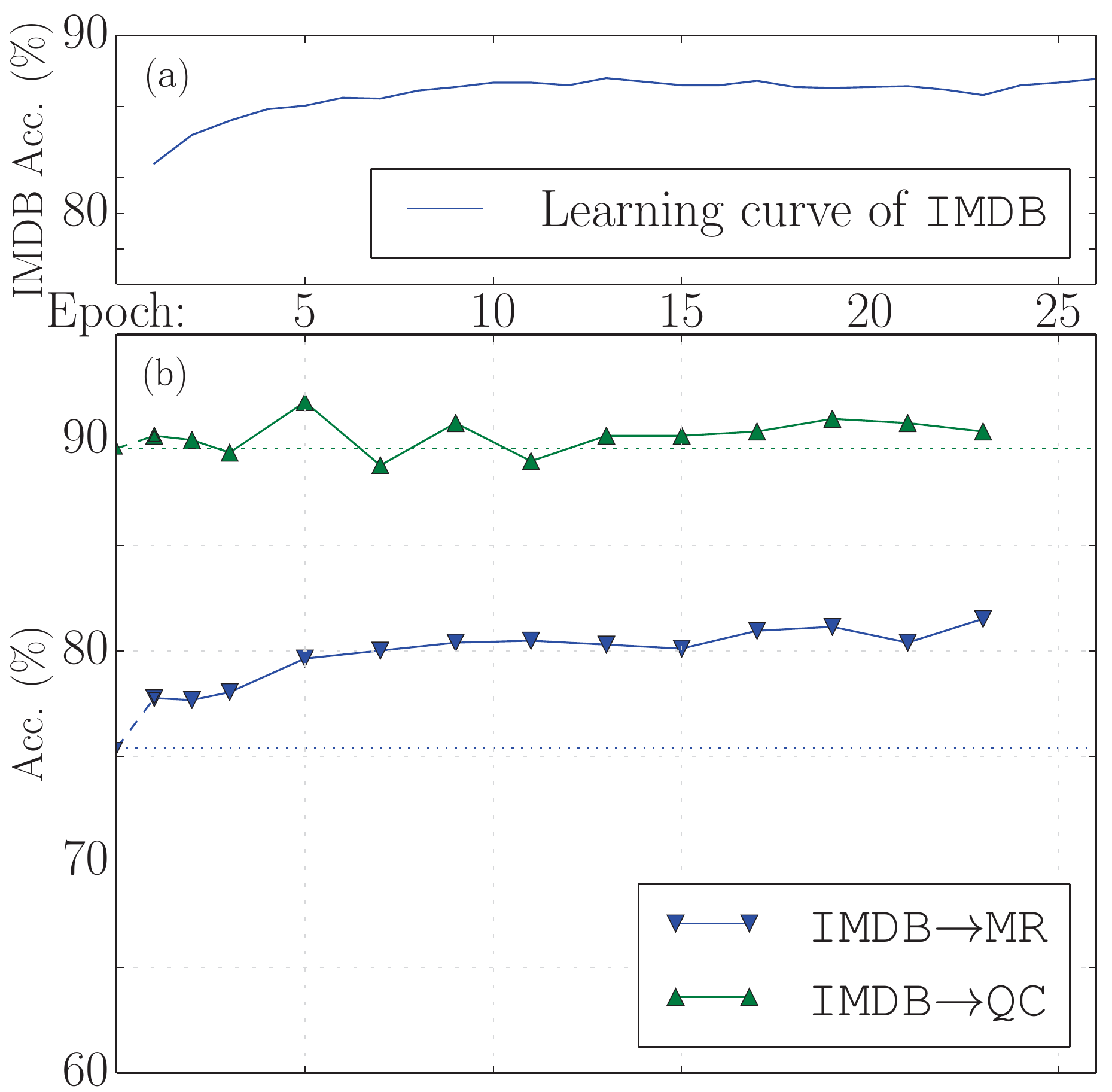}\\
\textbf{\quad Experiment~II}\\
\includegraphics[width=.42\textwidth]{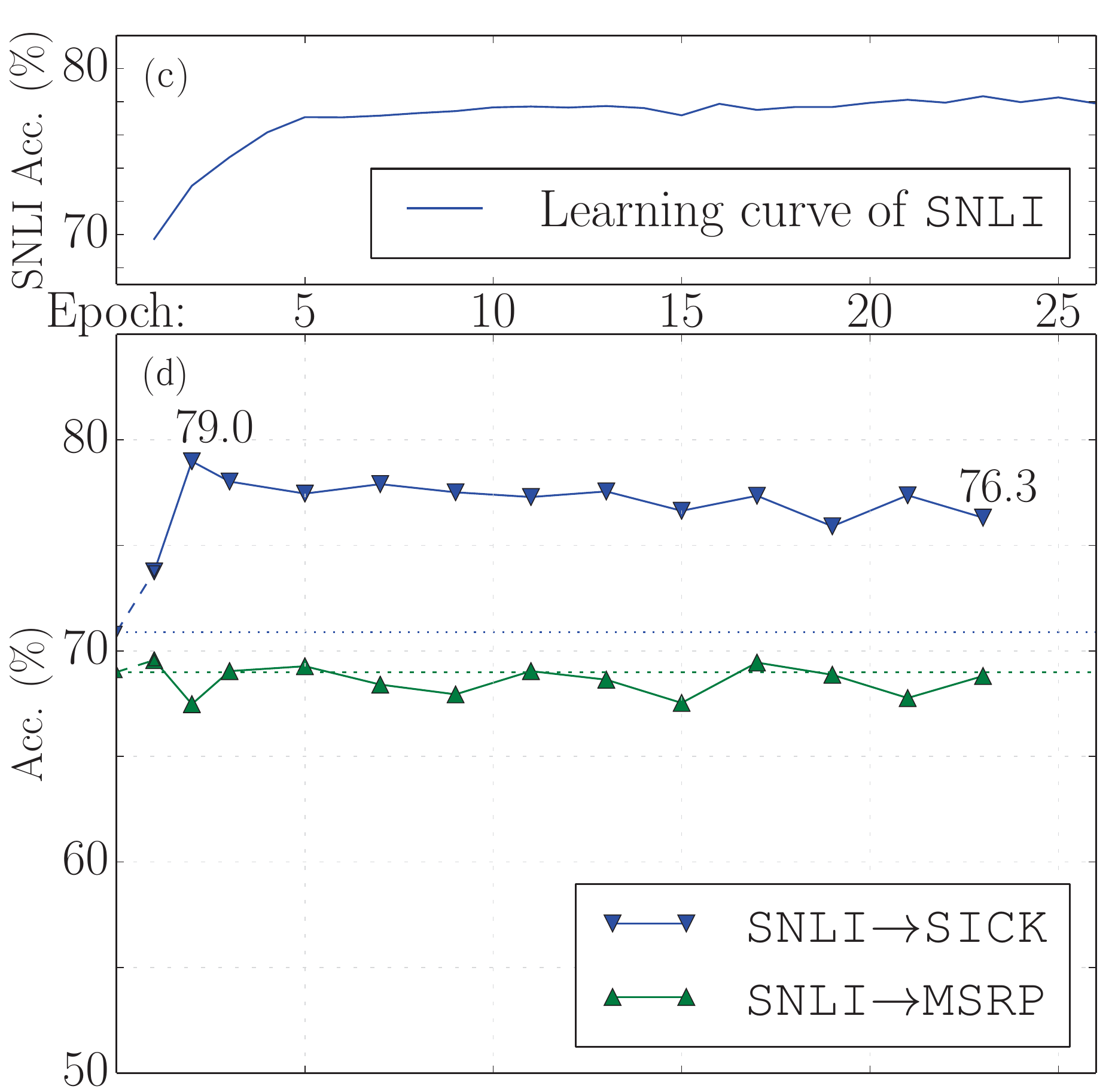}
\caption{(a) and (c): Learning curves of $\mathcal{S}$. (b) and (d): Accuracies of $\mathcal{T}$ when parameters are transferred at a certain epoch during the training of $\mathcal{S}$. Dotted lines refer to non-transfer, which can be equivalently viewed as transferring before training on $\mathcal{S}$, i.e., epoch $=0$. Note that the $x$-axis shares across different subplots.
}\label{fig:when}
\end{figure}

However, we may imagine that the parameters well-tuned to the source dataset may be too specific to it, i.e., the model overfits $\mathcal{S}$ and thus may underfit $\mathcal{T}$.
Another advantage of early transfer lies in computational concerns. If we manage to transfer model parameters after one or a few epochs on $\mathcal{S}$, we can save much time especially when $\mathcal{S}$ is large.

We therefore made efforts in studying when the neural model is ready to be transferred.
Figures~\ref{fig:when}a and \ref{fig:when}c plot the learning curves of the source tasks. The accuracy increases sharply from epochs 1--5; later, it reaches a plateau but is still growing slowly. 

We then transferred the parameters at different stages (epochs) of training to target tasks (also with the setting E\unlock H\unlock O$\Box$). Their accuracies are plotted in Figures~\ref{fig:when}b and~\ref{fig:when}d.

%Transferring to semantically different tasks at any epoch appears to be frustratingly ineffective.
%No matter how the model performs on $\mathcal{S}$, it can hardly fit $\mathcal{T}$ better than without transfer.
%The results rule out some undesirable factors (e.g., overfitting the source domain) that may cause the failure to transfer from $\mathcal{S}$ to $\mathcal{T}$ in Subsection~\ref{ss:performance}.

In \imdbmr, the source performance and transferring performance  align well.
The \snlisick\ experiment, however, produces interesting yet unexpected results. Using the second epoch of \snli's training yields the highest transfer performance on \sick, i.e., 78.98\%, when the \snli\ performance itself is comparatively low (72.65\% vs.~76.26\% at epoch 23). Later, the transfer performance decreases gradually by $\sim$2.7\%. %Although the degradation is not significant, the tendency is reasonably perceptible, showing that by fitting $\mathcal{S}$ too well, the parameters may be less effective for $\mathcal{T}$ more or less. We have our second additional finding as follows. 
%More evidence is needed in order to draw conclusions. Nevertheless, from Figure~\ref{fig:when}, we are reasonably safe to conclude
The results in these two experiments are inconsistent and lack explanation.
%\smallskip
%\noindent\fbox{
%\parbox{.45\textwidth}{
%When we apply INIT for transferring, only a few epochs over the source dataset may suffice to capture transferable knowledge, although the source task performance has not been optimal.
%}
%}

\section{MULT, and its Combination with INIT}\label{sec:MULT}

To answer RQ3, we investigate how multi-task learning performs in transferring knowledge, as well as the effect of the combination of MULT and INIT. In this section, we applied the setting: sharing embeddings and hidden layers (denoted as E\myheart H\myheart O$\Box$), analogous to E\unlock H\unlock O$\Box$\ in INIT. %Note that sharing all parameters  E\myheart H\myheart O\myheart\ is not applicable to \msrp\ because of different output objectives; thus we did not apply this setting. 
When combining MULT and INIT, we used the pretrained parameters of embeddings and hidden layers on $\mathcal{S}$ to initialize the multi-task training of $\mathcal{S}$ and $\mathcal{T}$, visually represented by E\unlock\myheart H\unlock\myheart O$\Box$.

\begin{figure}[!t]
\centering
\textbf{\quad Experiment~I}\\
\includegraphics[width=.45\textwidth]{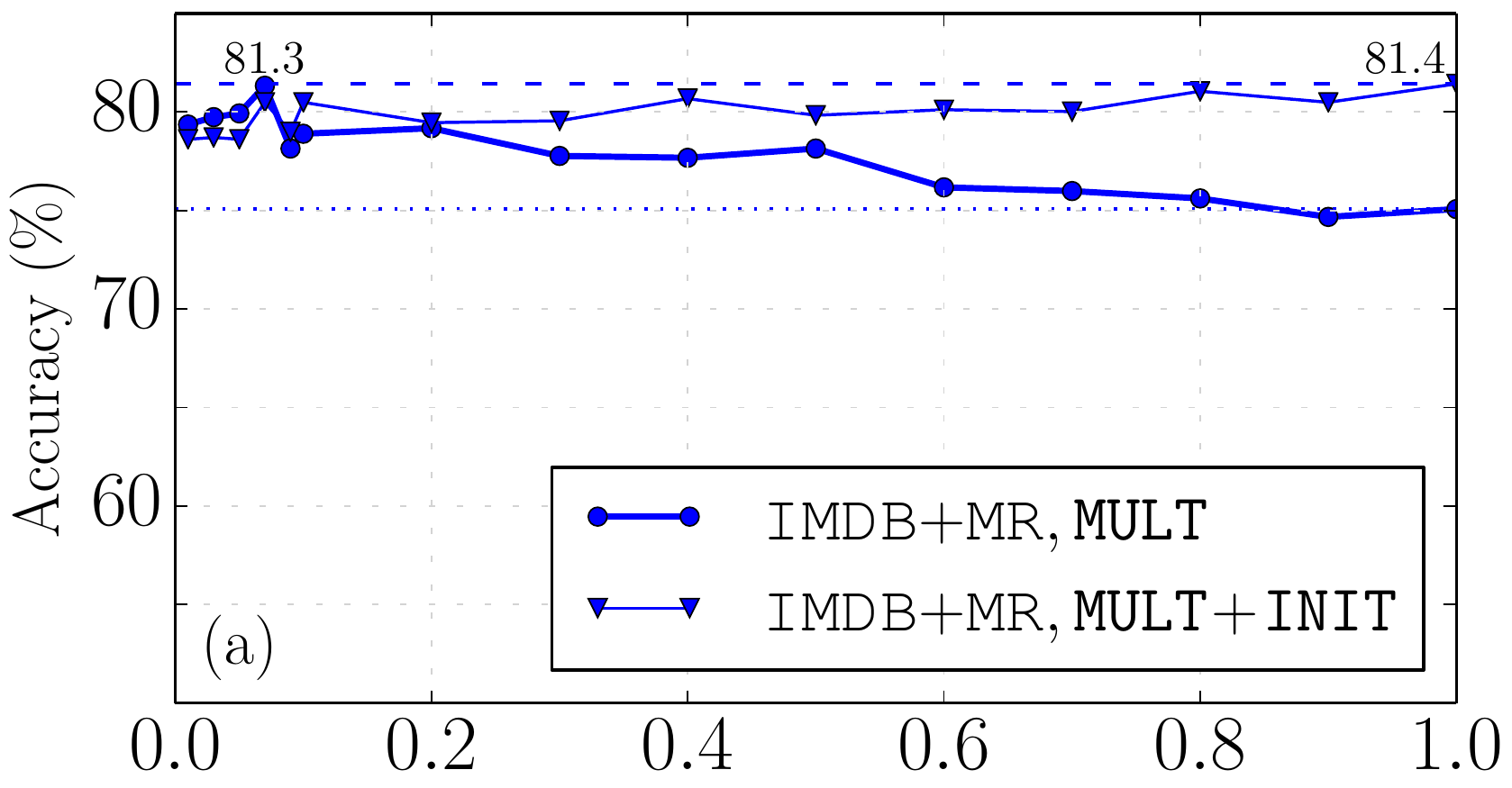}
\includegraphics[width=.45\textwidth]{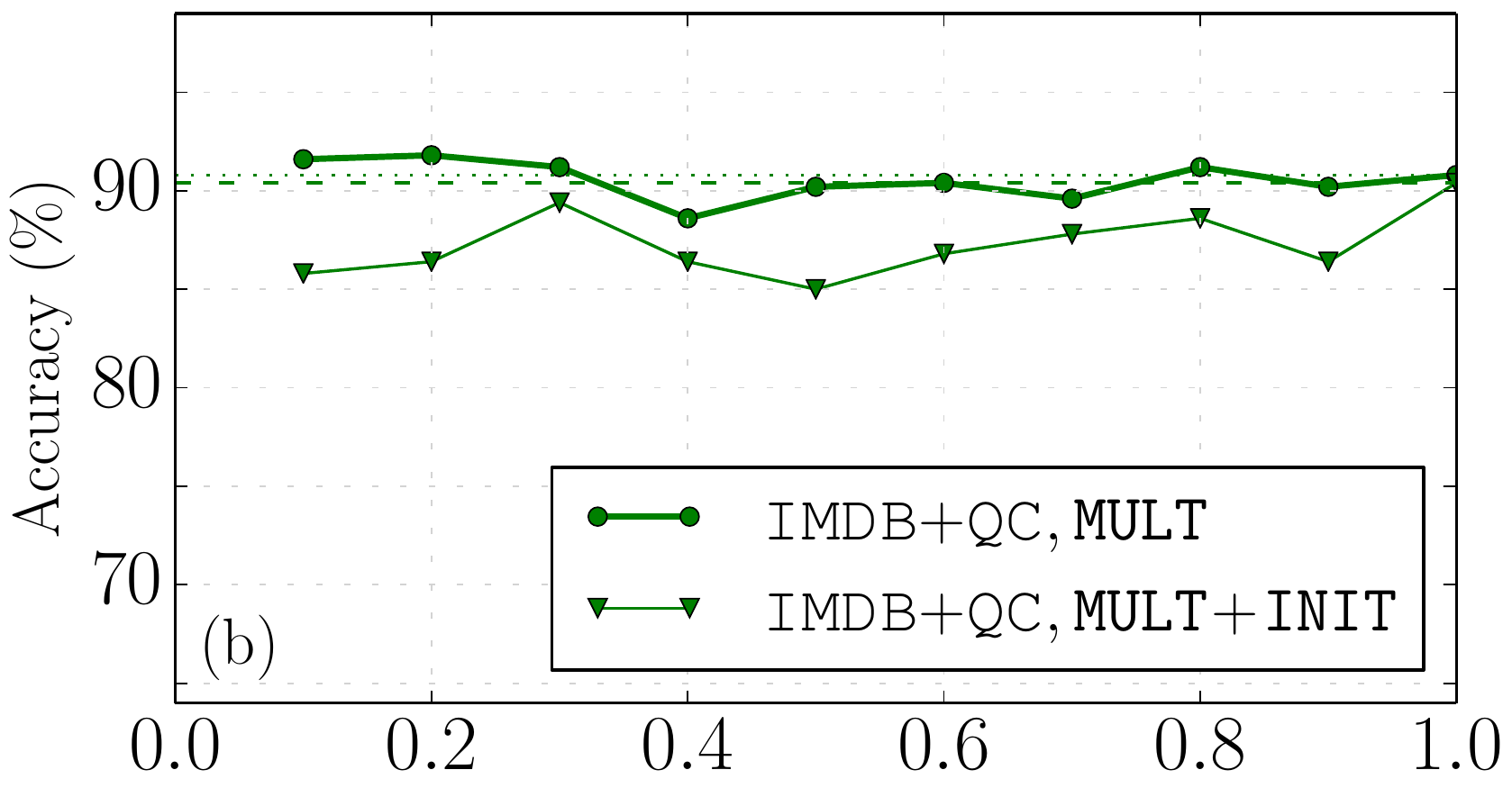}
\textbf{\quad Experiment~II}\\
\includegraphics[width=.45\textwidth]{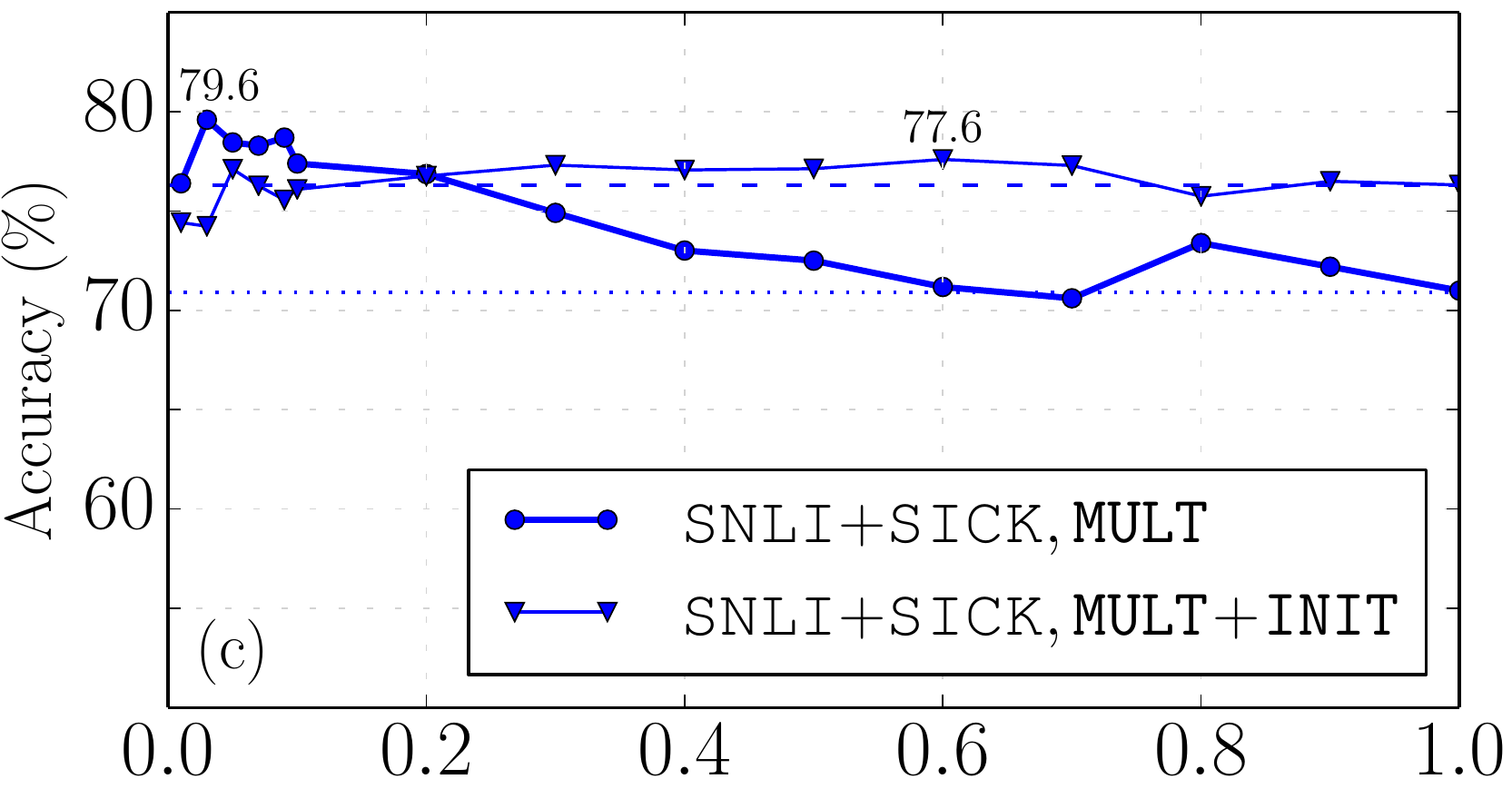}
\includegraphics[width=.45\textwidth]{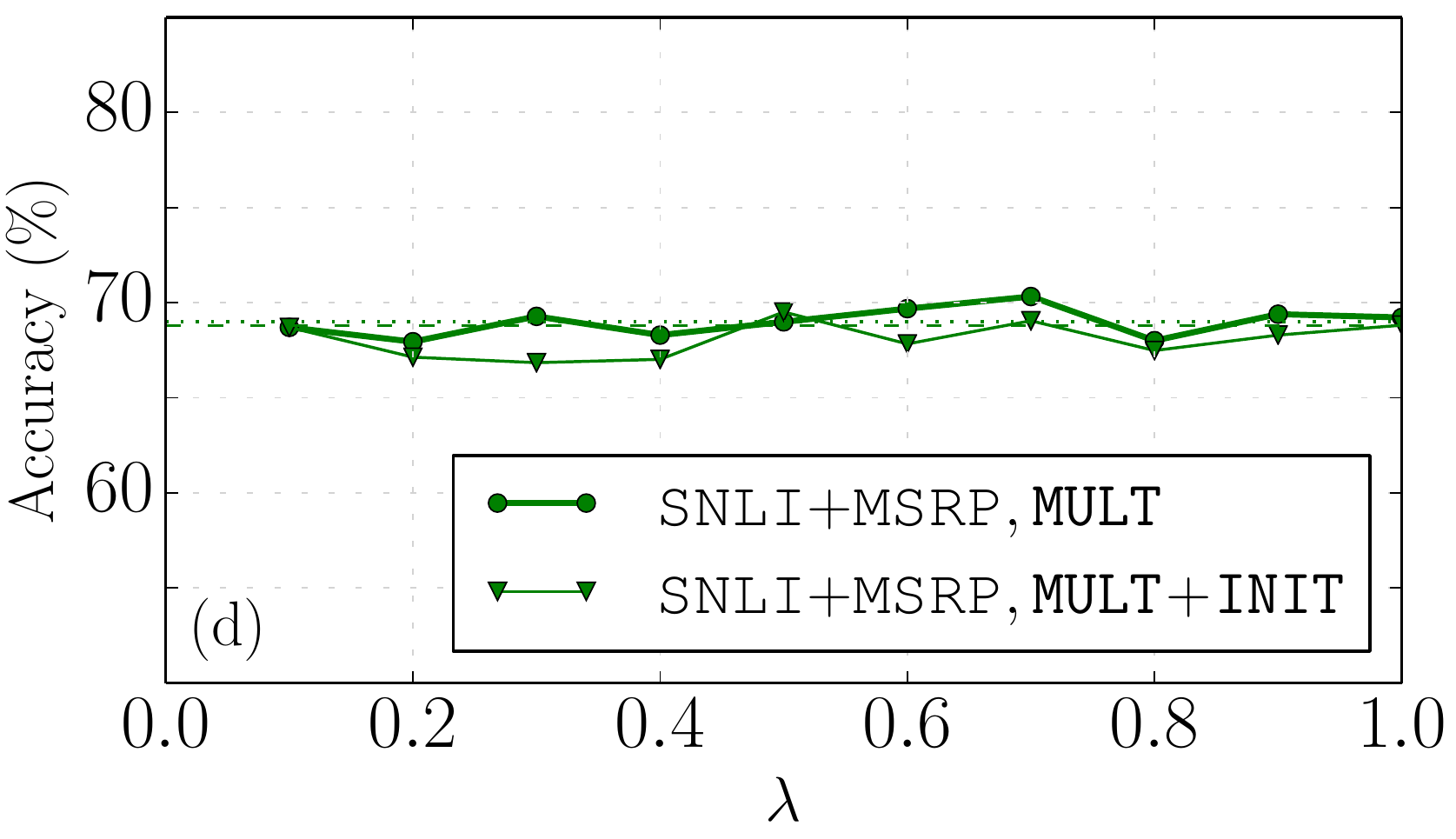}
\caption{Results of MULT and MULT+INIT, where we share word embeddings and hidden layers. Dotted lines are the non-transfer setting; dashed lines are the INIT setting E\unlock H\unlock O$\Box$, transferred at the peak performance of \imdb\ and \snli.}
\label{fig:MULT}
\end{figure}

In both MULT and MULT+INIT, we had a hyperparameter $\lambda\in(0,1)$ balancing the source and target tasks (defined in Section~\ref{sec:transfer}). $\lambda$ was tuned with a granularity of 0.1. As a friendly reminder, $\lambda=1$ refers to using $\mathcal{T}$ only; $\lambda=0$ refers to using $\mathcal{S}$ only.
After finding that a small $\lambda$ yields high performance of MULT in the \imdb+\mr\ and \snli+\sick\ experiments (thick blue lines in Figures~\ref{fig:MULT}a and \ref{fig:MULT}c), we further tuned the $\lambda$ from 0.01 to 0.09 with a fine-grained granularity of 0.02.

The results are shown in Figure~\ref{fig:MULT}. From the green curves in the 2nd and 4th subplots, we see MULT (with or without INIT) does not improve the accuracy of target tasks (\qc\ and \msrp); the inability to transfer is cross-checked by the INIT method in Section~\ref{sec:INIT}. For \mr\ and \sick, on the other hand, transferability of the neural model is also consistently positive (blue curves in Figures~\ref{fig:MULT}a and~\ref{fig:MULT}c), 
supporting our conclusion to RQ1 that neural transfer learning in NLP depends largely on how similar in semantics the source and target datasets are.

Moreover, we see that the peak performance of MULT is slightly lower than INIT in Experiment I (Figure~\ref{fig:MULT}a), but higher in Experiment II (Figure~\ref{fig:MULT}c); they are in the same ballpark.
%When $\lambda$ is too small (say, 0.01), the performance drops, as it fits $\mathcal S$ too much and thus underfits $\mathcal T$.

In MULT+INIT (E\unlock\myheart H\unlock\myheart O$\Box$), the transfer performance of MULT+INIT remains high for different values of $\lambda$. Because the parameters given by INIT have already conveyed sufficient information about the source task, MULT+INIT consistently outperforms non-transferring by a large margin. Its peak performance, however, is not higher than MULT or INIT. 
In summary, we answer our RQ3 as follows:
in our experiments, MULT and INIT are generally comparable; we do not obtain further gain by combining MULT and INIT.
%\footnote{
%For aesthetic purposes, the main results that have been boxed in Section~\ref{sec:intro} are not boxed again.}

\section{Concluding Remarks}\label{sec:conclusion}

In this paper, we addressed the problem of transfer learning in neural network-based NLP applications. We conducted two series of experiments on six datasets, showing that the transferability of neural NLP models depends largely on the semantic relatedness of the source and target tasks, which is different from other domains like image processing. We analyzed the behavior of different neural layers. We also experimented with two transfer methods: parameter initialization (INIT) and multi-task learning (MULT). Besides, we reported two additional studies in Sections~\ref{ss:learningRate} and~\ref{ss:when} (not repeated here).
Our paper provides insight on the transferability of neural NLP models; the results also help to better understand neural features in general.

\textbf{How transferable are the conclusions in this paper?}
%Although we had in total more than 5000 separate runs of experiments in this paper, 
We have to concede that empirical studies are subject to a variety of factors (e.g., models, tasks, datasets), and that conclusions may vary in different scenarios. 
In our paper, we have tested all results on two groups of experiments involving 6 datasets and 2 neural models (CNN and LSTM-RNN). Both models and tasks are widely studied in the literature, and not chosen deliberately. Results are mostly consistent (except Section~\ref{ss:when}).
Along with analyzing our own experimental data, we have also collected related results in previous studies, serving as additional evidence in answering our research questions. Therefore, we think the generality of this work is fair and that the conclusions can be generalized to similar scenarios.

\textbf{Future work.} Our work also points out some future directions of research. For example, we would like to analyze the effect of different MULT strategies. More efforts are also needed in developing an effective yet robust method for multi-task learning.

\section*{Acknowledgments}
We thank all reviewers for their constructive comments, Sam Bowman for helpful suggestion, and Vicky Li for discussion on the manuscript. This research is supported by the National Basic Research Program of China (the 973 Program) under Grant No.~2015CB352201 and the National Natural Science Foundation of China under Grant Nos.~61232015, 91318301, 61421091, 61225007, and 61502014.

\bibliographystyle{acl2016}
\bibliography{transfer}

\end{document}